\pdfoutput=1
\documentclass[10pt,twocolumn,letterpaper]{article}
\usepackage[pagenumbers]{cvpr}

\usepackage{microtype}

\usepackage{booktabs}
\usepackage{flafter}

\usepackage{makecell}
\usepackage{multirow}

\definecolor{cvprblue}{rgb}{0.21,0.49,0.74}
\usepackage[pagebackref,breaklinks,colorlinks,allcolors=cvprblue]{hyperref}

\def\confName{CVPR}
\def\confYear{2027}

\title{Making the World React Realistically: A Closed-Loop Interactive World Model for Autonomous Driving}

\author{
Jia Hu$^{1}$ \hspace{0.75em}Yang Chang$^{1}$ \hspace{0.75em}Panqu Wang$^{2}$ \hspace{0.75em}Zhijun Fu$^{3}$ \hspace{0.75em}Chang Liu$^{4,5}$ \hspace{0.75em}Changjian Yu$^{1}$ \hspace{0.75em}Jintao Lai$^{1,*}$\\[0.6em]
$^{1}$Tongji University \quad $^{2}$ZERON \quad $^{3}$Zhengzhou University of Light Industry\\
$^{4}$Budapest University of Technology and Economics \quad $^{5}$HUN-REN SZTAKI\\[0.3em]
{\small $^{*}$Corresponding author: {\tt Jintao\_Lai@tongji.edu.cn}}
}

\begin{document}
\maketitle
\begin{abstract}
World models (WMs) generate driving scenarios beyond recorded logs, enabling the training and evaluation of autonomous driving systems in unseen interactions. However, existing WMs struggle to ensure both behavioral realism and visual fidelity, since they overlook the relationship between vehicle interactions and the ego visual observations. We propose a closed-loop interactive WM that couples behavior modeling and visual rendering. First, imitation-learned background-vehicle (BV) policies are fine-tuned with multi-agent reinforcement learning in the closed loop against an end-to-end ego planner that reacts to the rendered images, so that their interaction behaviors match those in real logs. Second, the viewpoints exposed by these interactions guide a per-Gaussian diagnosis of each 3D Gaussian vehicle asset, which attenuates view-dependent colors on under-observed Gaussians and fills never-observed surfaces with generated Gaussians, while preserving the observed appearance. Third, signal lights are rendered according to the driving intentions of the vehicles, by an editor that learns the light states unobserved for a vehicle through counterfactual fine-tuning with a light-state classifier. On long-tail closed-loop scenarios, our method reduces the divergence from real-world interaction behaviors by $85\%$ and improves novel-view vehicle PSNR by $2.0$\,dB over the state-of-the-art WM. The significance of BV light signaling in WMs is shown, as the rendered signal lights conveying BV driving intentions indirectly impact ego driving behaviors in a quarter of the scenarios.
\end{abstract}
\section{Introduction}
\label{sec:intro}

Autonomous driving has advanced rapidly, driven in large part by the availability of real-world driving datasets.
Large-scale logs~\cite{caesar2021nuplan, dauner2024navsim} provide synchronized images, trajectories, and maps, serving as the main source of both training and evaluation for driving policies.

However, driving datasets only record real-world scenarios and cannot generate unseen ones, causing autonomous driving systems to perform poorly in unseen scenarios.
Each log in the dataset captures a single realization of traffic, and safety-critical events such as sudden cut-ins or hard braking are rare in it~\cite{wang2023testing,hu2025accelerated,xu2023critical,chen2026prior}.

World models (WMs) offer a data-driven approach to generate diverse driving scenarios beyond the recorded ones.
According to what responds to the ego vehicle's actions, existing driving WMs fall into three categories~\cite{xu2026foresight}.
\emph{Replay-constrained} WMs only render observations from altered viewpoints, while the ego vehicle and background vehicles (BVs) follow prescribed trajectories; \emph{ego-responsive} WMs move the ego vehicle by its own actions while BVs still follow prescribed trajectories; and \emph{traffic-reactive} WMs further let BVs respond to the ego vehicle.

\emph{Replay-constrained} WMs cannot embed the ego vehicle into their generated driving scenarios, making it impossible to evaluate the safety of the ego vehicle in such scenarios.
Models of this type, such as SplatAD~\cite{hess2025splatad} and NeuRAD~\cite{tonderski2024neurad}, cannot move the ego vehicle by its own actions.

\emph{Ego-responsive} WMs~\cite{gao2024vista, wang2024driving} allow the ego vehicle to be embedded into generated driving scenarios and take actions, but BVs in the scenarios cannot respond to the ego vehicle. 
Hence, they fail to test the risks of the ego vehicle when interacting with BVs.

\emph{Traffic-reactive} WMs~\cite{yang2024drivearena,you2024bench2drive} allow interactions between the ego vehicle and BVs.
However, the rendering performance of the driving scenarios will severely degrade on viewpoints falling out of the recorded trajectory.
This is because logs observe the scenarios only from viewpoints along the recorded ego trajectory, so views exposed by interactions, such as unseen vehicle sides, lack supervision and are rendered with artifacts.
Some works~\cite{zhou2024hugsim,jiang2025realengine} instead insert vehicle assets reconstructed from external 360-degree datasets~\cite{du20243drealcar}, but these assets alter the appearance (\eg, model and color) of the logged vehicles.

The state-of-the-art (SOTA) WM improves visual fidelity by combining novel-view synthesis with reactive BV behaviors.
World Engine~\cite{li2026worldengine} reconstructs driving scenarios with 3D Gaussian Splatting (3DGS)~\cite{kerbl20233dgs}, completes unseen vehicle views with Asset Harvester~\cite{assetharvester2026}, and drives BVs with a diffusion-based behavior model that reacts to the ego vehicle.

However, it remains challenging for existing methods to ensure both BV behavioral realism and visual fidelity, since they overlook the relationship between vehicle interactions and visual observations.
In a closed loop, a BV interaction changes what the ego vehicle observes, and the ego reaction to this rendered observation in turn changes how BVs behave through \emph{visual feedback}—a coupling that existing WMs ignore by handling behavior modeling and visual rendering separately.
This challenge manifests in three aspects.

\textbf{First}, existing methods may generate unrealistic BV interaction behaviors as they fail to learn how BV interactions change when visual observations fluctuate.
BV behavior models in existing WMs~\cite{yang2024drivearena,li2026worldengine} are built on vehicle states alone. In a WM, however, the ego vehicle reacts to rendered images rather than the states, so BVs face ego reactions they are not built for, and the distribution of their interaction features, such as time-to-collision, drifts away from that of real-world driving logs~\cite{ross2011dagger,zhang2025catk}.

\textbf{Second}, existing methods struggle to reproduce high-fidelity novel views induced by vehicle interactions, as they mistakenly replace the well-fitted regions of 3D Gaussian assets.
Interactions such as cut-ins may expose vehicle sides absent from the log.
To render them, existing pipelines~\cite{assetharvester2026,li2026worldengine} synthesize images of the vehicle from all directions and rebuild a new 3D Gaussian asset from them, which replaces the entire original asset. As a result, even the regions well fitted to real images are replaced by generated ones.

\textbf{Third}, existing methods fail to generate realistic vehicle signal lights, as they do not model how these signals convey interactive driving intentions.
Most reconstruction-based WMs~\cite{chen2025mtgs,li2026worldengine} bake signal light appearance into static colors, and NuRec~\cite{nurec2026} can only replay the logged signal light states.
Hence, when a BV takes actions different from the log, \eg, braking or changing lanes in response to the ego vehicle, its signal lights cannot change accordingly, although drivers and perception models rely on such cues~\cite{kahane1989chmsl,rasouli2019deepsignals}.

In this paper, we propose a closed-loop interactive world model that couples behavior modeling and visual rendering to address these challenges.
It features \textbf{(1) human-like vehicle interaction behaviors, (2) enhanced fidelity of novel views exposed by vehicle interactions, and (3) realistic vehicle light signaling triggered by interactive driving intentions}.
Our contributions are as follows:
\begin{itemize}
    \item \textbf{Visual-feedback-driven interaction learning for behavioral realism.} A style-conditioned mixture-of-experts policy learns diverse behaviors by imitation learning (IL), and is refined in the closed loop with visual feedback and calibrated by multi-agent reinforcement learning (MARL) toward real-world interaction feature distributions.
    \item \textbf{Interaction-guided selective Gaussian asset enhancement.} Viewpoints exposed by vehicle interactions in the closed loop guide a per-Gaussian diagnosis. Only the regions diagnosed as deficient are repaired with Gaussians generated by multi-view diffusion, preserving the regions well fitted to real images.
    \item \textbf{Intention-triggered vehicle light signaling via counterfactual fine-tuning.} 
    A BV's signal lights follow its driving intentions, but some intentions are absent from the log. To render their corresponding light states, we adopt counterfactual fine-tuning: a diffusion-based editor generates the light states of intentions that did not occur in fact, and a light-state discriminator supervises whether the outputs match the intentions.
\end{itemize}

\section{Related Work}
\label{sec:relatedwork}

\noindent\textbf{BV interaction behaviors in WMs.}
Closed-loop driving WMs require BVs to evolve over time and, ideally, to respond to the ego vehicle~\cite{hu2024simulation,hu2025simulation}.
The simplest BV model~\cite{gao2024vista,ljungbergh2024neurocap} replays logged trajectories, which cannot react once the ego deviates from the log.
Rule-based models~\cite{zhou2024hugsim,you2024bench2drive,yang2024drivearena} in WMs respond to the ego but cannot reproduce the diverse behaviors of human drivers.
Data-driven models~\cite{wang2026behaviorworldgen,jiang2025realengine,li2026omninwm,li2026worldengine} learn diverse behaviors from large-scale logs, but none of them learns how BV interactions change with the ego's visual observations; we instead fine-tune BV policies with visual feedback in the closed loop.

\noindent\textbf{Novel view synthesis in WMs.}
Reconstruction-based WMs~\cite{yan2024street,tonderski2024neurad,hess2025splatad} built on NeRF~\cite{mildenhall2021nerf} or 3DGS~\cite{kerbl20233dgs} are faithful near the recorded trajectory but degrade as the viewpoint departs from it.
Diffusion-based enhancers~\cite{zhao2025recondreamer,wu2025difix3d,diffusionharmonizer2026} restore degraded renderings, but hallucinate content where no geometry exists and cannot recover never-observed vehicle sides.
Asset-level solutions~\cite{zhou2024hugsim,jiang2025realengine} replace scene vehicles with external 3D assets, at the cost of losing the appearance of the original vehicles.
Asset Harvester~\cite{assetharvester2026} instead generates appearance-preserving assets from sparse observations, and World Engine~\cite{li2026worldengine} integrates it with MTGS~\cite{chen2025mtgs}, but this pipeline replaces the well-fitted regions of native assets.
We instead diagnose the deficient regions that interactions expose and repair only them, preserving native appearance elsewhere.

\noindent\textbf{Vehicle light signaling in WMs.}
Vehicle signal lights are critical visual cues for communicating driving intentions.
Brake lights reduce rear-end collisions~\cite{kahane1989chmsl,farmer1996chmsl} and reaction times~\cite{li2014flashing,xue2018perceptual}, and end-to-end planners change their decisions when light states are edited~\cite{Jacob2022STEEX,Zemni2023OCTET,semanticshapley2025}.
However, reconstruction-based WMs~\cite{chen2025mtgs,yan2024street,zhou2024hugsim,li2026worldengine} bake signal lights into static colors, generative WMs~\cite{hu2025gaia2,gao2024vista,yan2026causaldrive} do not expose light states as a controllable signal, and NuRec~\cite{nurec2026} only replays logged light transitions.
General instruction-based image editors~\cite{brooks2023instructpix2pix,zhang2023magicbrush,labs2025flux} can paint signal lights frame by frame, but ignore 3D lamp locations and driving intentions, alter the vehicle body, and flicker across frames.
To our knowledge, we are the first to bring intention-triggered light signaling into a closed-loop interactive WM.
\section{Methodology}
\label{sec:method}
\begin{figure*}[!t]
    \centering
    \includegraphics[width=0.95\textwidth]{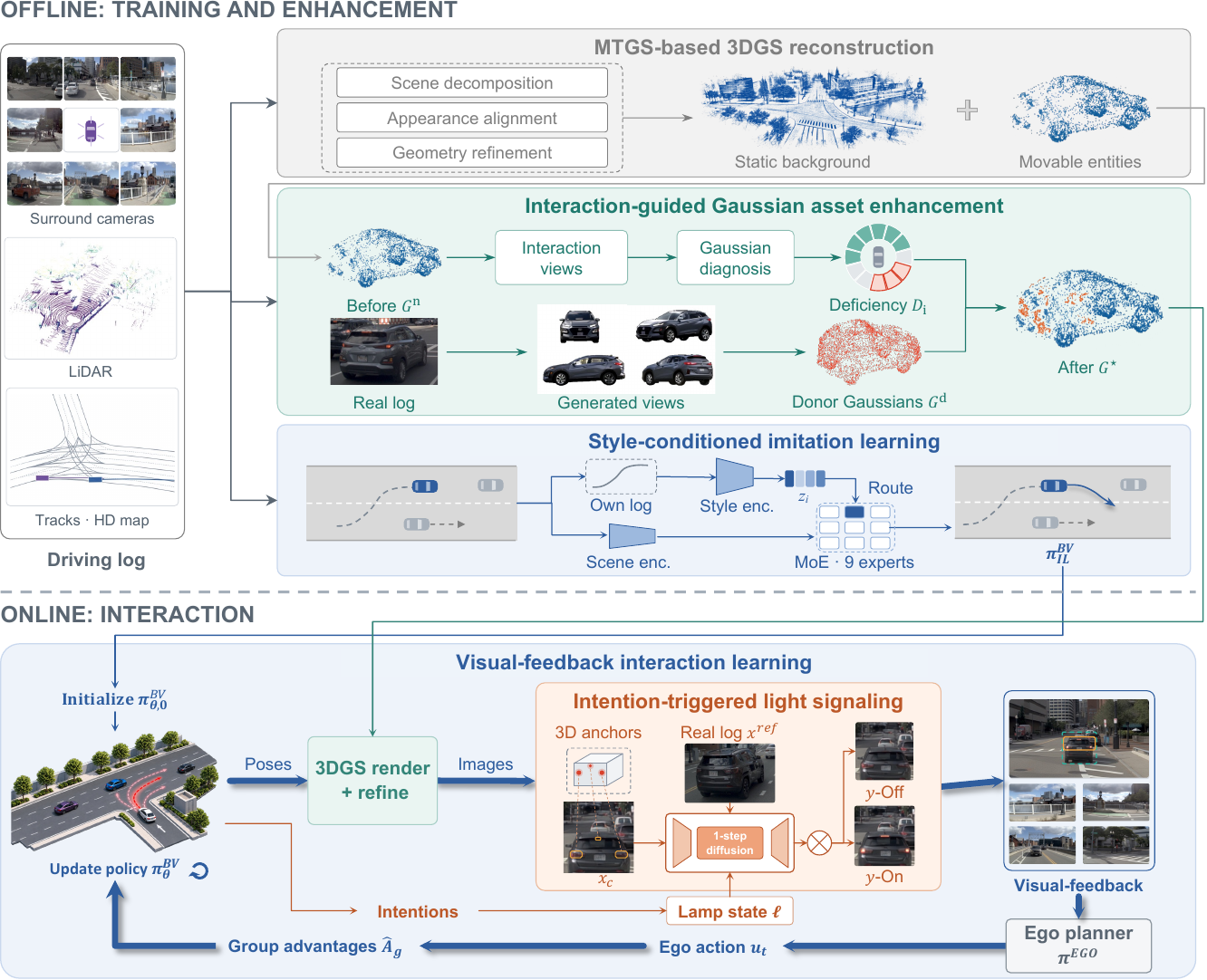}
    \caption{Overview of our world model. Offline: driving logs support 3DGS reconstruction, interaction-guided asset enhancement, and style-conditioned IL. Online: the BV policy is fine-tuned with the ego planner on renderings with intention-triggered signal lights.}
    \label{fig:overview}
\end{figure*}

As shown in \cref{fig:overview}, our world model builds 3D Gaussian assets and initial BV policies offline (\cref{sec:asset_enhancement,sec:imitation_learning}), and then renders intention-triggered signal lights and refines BV policies in the closed loop online (\cref{sec:light_signaling,sec:interaction_learning}).
Each scene is first reconstructed with MTGS~\cite{chen2025mtgs} from surround-view cameras, LiDAR, and tracked boxes into a static background and movable vehicle assets, where each vehicle is represented by Gaussians in its own box frame so that it can be re-posed and rendered along its simulated trajectory.

\subsection{Interaction-guided Gaussian Asset Enhancement}
\label{sec:asset_enhancement}
A logged vehicle is seen from only a few directions, so its native asset $G^{\mathrm{n}}=\{g_i\}$ is unreliable from the others.

\noindent\textbf{Interaction-aware Gaussian diagnosis.}
For Gaussian $g_i$ in training view $t$, let $\phi_{i,t}$ be the azimuth of the camera seen from the Gaussian in the box frame ($0$ at the heading), and $c_{i,t}=\sum_p c_{i,t}(p)$ its total blending weight, where $c_{i,t}(p)$ is its opacity times transmittance at pixel $p$~\cite{kerbl20233dgs}.
Its support at azimuth $\phi$ accumulates nearby observations:
\begin{equation}
    S_i(\phi)=\sum_t c_{i,t}\,\kappa(\phi-\phi_{i,t})+\lambda\sum_t c_{\bar{i},t}\,\kappa(\phi+\phi_{\bar{i},t}),
\end{equation}
where $\kappa$ is a periodic Gaussian kernel and $g_{\bar{i}}$ is the left--right mirrored copy that MTGS keeps of $g_i$, whose observations from $\phi$ constrain $g_i$ from $-\phi$ with a discount $\lambda<1$.
Rendering the asset from probe views $v$, we define the deficiency of $g_i$ as the unsupported share of its contribution:
\begin{equation}
    D_i=\frac{\sum_v w_v\,c_{i,v}\,\bigl(1-\tilde{S}_i(\phi_{i,v})\bigr)}{\sum_v w_v\,c_{i,v}},\quad \tilde{S}_i=1-e^{-S_i/s_0},
\end{equation}
where $s_0$ is a saturation scale.
Probe views combine uniform $360^\circ$ views with higher-weighted interaction-critical views sampled from ego-camera directions in closed-loop rollouts (\eg, cut-ins), so $D_i$ is high for Gaussians that interactions expose but the log does not constrain.

\noindent\textbf{Repairing with donor Gaussians.}
A native Gaussian with high $D_i$ shows extrapolated colors from unsupported directions, so we scale its view-dependent spherical-harmonic coefficients by $1-D_i$ and keep its base color, giving $\hat{g}_i$.
Moreover, never-observed surfaces contain no Gaussians at all; such holes are invisible to $D_i$ and are filled from a donor asset $G^{\mathrm{d}}$, generated by Asset Harvester~\cite{assetharvester2026} via multi-view synthesis from the vehicle's image crops in the log and registered to $G^{\mathrm{n}}$.
For each donor Gaussian $g$, the share of its probe-view contribution that falls on pixels left empty by $G^{\mathrm{n}}$ is
\begin{equation}
    u_g=\frac{\sum_v w_v\sum_p c_{g,v}(p)\,\bigl(1-A^{\mathrm{n}}_v(p)\bigr)}{\sum_v w_v\,c_{g,v}},
\end{equation}
where $A^{\mathrm{n}}_v(p)$ is the accumulated opacity of $G^{\mathrm{n}}$ at pixel $p$.
The repaired asset $G^{\star}=\{\hat{g}_i\}\cup\{g\in G^{\mathrm{d}}:u_g>u_0\}$ is then fine-tuned on training images.

\noindent\textbf{Image-space refinement.}
The rendering of $G^{\star}$ is finally refined in image space.
(i)~\emph{Color calibration}: a per-camera, per-channel gain and bias fitted on training images is applied to the rendering of $G^{\star}$, giving $x$.
(ii)~\emph{Bounded fusion}: 
two corrections of $x$ are available, the output $\tilde{x}$ of a diffusion harmonizer~\cite{diffusionharmonizer2026}, which looks more realistic but may hallucinate, and the donor residual $\delta^{\mathrm{d}}=m\odot(x^{\mathrm{d}}-x)$, where $x^{\mathrm{d}}$ is rendered with $G^{\mathrm{d}}$ and $m$ marks pixels still missed by $G^{\star}$ but covered by the donor.
A lightweight U-Net $f_\psi$ predicts how much of each to accept per pixel:
\begin{equation}
    \hat{x}=x+\sigma(h_1)\,(\tilde{x}-x)+\sigma(h_2)\,\delta^{\mathrm{d}}+\epsilon\tanh(h_3),
\end{equation}
where $[h_1,h_2,h_3]=f_\psi(x,\tilde{x},\delta^{\mathrm{d}},m)$, $\sigma$ is the sigmoid, and $\epsilon$ bounds the free residual; $f_\psi$ is trained on training images.

\subsection{Style-conditioned Imitation Learning}
\label{sec:imitation_learning}
\begin{figure*}[!t]
    \centering
    \includegraphics[width=0.9\textwidth]{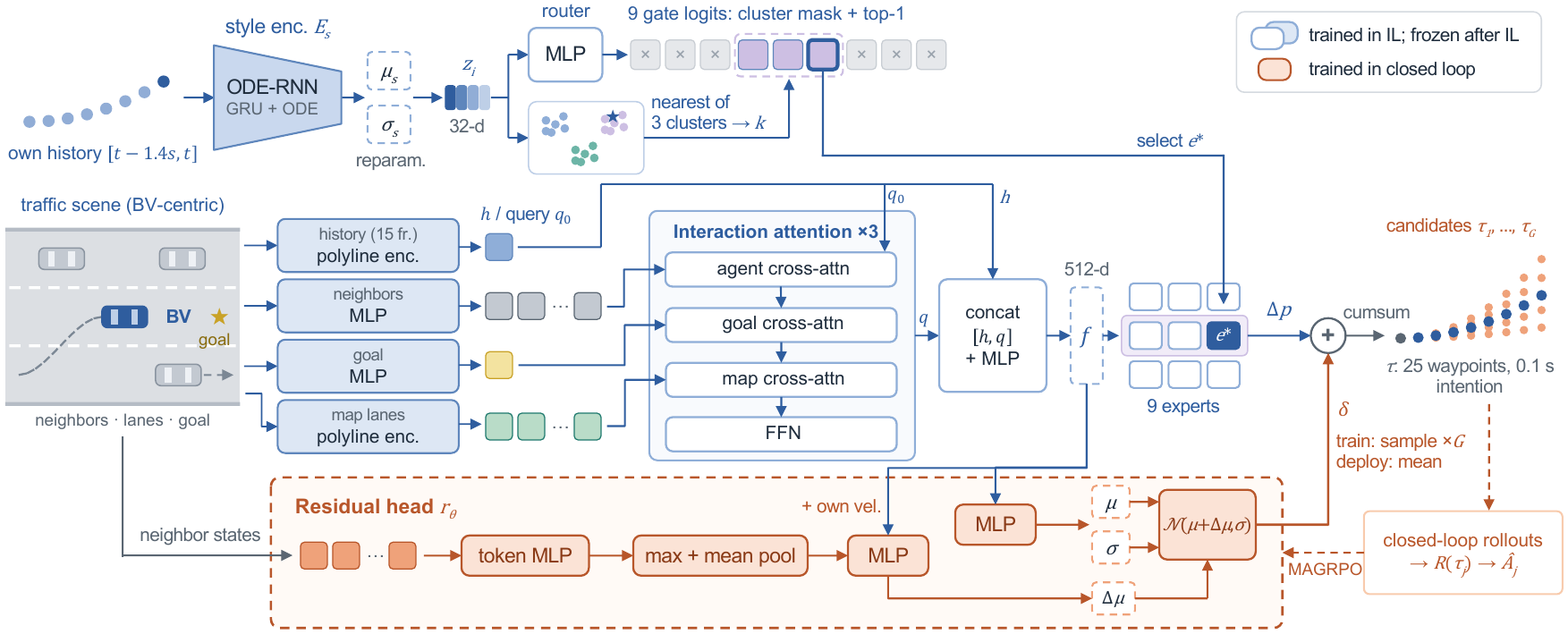}
    \caption{Architecture of the BV behavior model. The style latent $z_i$ routes to one expert $e^{*}$ in its style cluster, which maps the scene feature $f$ to displacements $\Delta p$. After IL, the policy (blue) is frozen and the residual head $r_\theta$ (orange), trained with MARL, adds an offset $\delta$.}
    \label{fig:bv_policy}
\end{figure*}

To make BVs drive like humans, we initialize BV policies by IL on recorded trajectories (\cref{fig:bv_policy}).
A pretrained style encoder $E_s$ encodes the BV's recent motion into a style latent $z_i$ assigned to one of $3$ style clusters~\cite{chen2026driving,lai2026style2risk}, and a scene encoder yields a scene feature $f$ from its history, neighbors, goal and map.
A router selects one of the $3$ experts in the cluster, denoted $e^{*}$, which decodes $f$ into $K$ per-step displacements $\Delta p=\{\Delta p_k\}_{k=1}^{K}$ accumulated into a plan $\tau$.
The policy $\pi^{\mathrm{BV}}_{\mathrm{IL}}$ is trained with
\begin{equation}
    \mathcal{L}_{\mathrm{IL}}=\sum_{k=1}^{K}\bigl\|\Delta p_k-\Delta p^{\mathrm{gt}}_k\bigr\|^2+\lambda_{\mathrm{s}}\mathcal{L}_{\mathrm{smooth}}+\lambda_{\mathrm{r}}\mathcal{L}_{\mathrm{road}}+\lambda_{\mathrm{c}}\mathcal{L}_{\mathrm{safe}},
\end{equation}
where the first term regresses the logged displacements, $\mathcal{L}_{\mathrm{smooth}}$ penalizes accelerations along $\tau$, and $\mathcal{L}_{\mathrm{road}}$ and $\mathcal{L}_{\mathrm{safe}}$ penalize leaving the drivable area and entering a safety margin around neighbors.

\subsection{Intention-triggered Vehicle Light Signaling}
\label{sec:light_signaling}
Reconstructed assets bake signal lights into static colors, so BVs cannot visually signal their intentions.

\noindent\textbf{Intention-driven signal light states.}
Signal light states follow each BV's own plan $\tau$ and intention.
A state machine activates the brake light when the planned deceleration exceeds a threshold; turn signals are activated for the side of a lane-changing intention ahead of the visible lateral motion and blink with a $1$\,s period until the lane change finishes.
This yields a state $\ell\in\{0,1\}$ for each signal light of each BV at each step.

\noindent\textbf{3D lamp anchoring.}
Per-frame 2D lamp detection is temporally unstable on rendered vehicles, so we anchor lamps in 3D on each asset: lamp masks segmented in the vehicle's log images are back-projected into the box frame with the rendered asset depth, keeping voxels hit from multiple views, and unseen sides are mirrored left--right.
At render time, the anchors are projected with the current camera and vehicle pose into a lamp map $A$ and its mask $M$.

\noindent\textbf{Light-state editing with counterfactual fine-tuning.}
A signal light is the effect of a driving intention, but the log shows each vehicle only under its factual intentions and thus in few light states; its appearance under another intention (\eg, braking instead of cruising) is a counterfactual without a paired image.
We fine-tune the one-step diffusion model Difix~\cite{wu2025difix3d} with LoRA~\cite{hu2022lora} into an editor $\mathcal{E}$ conditioned on the rendered vehicle crop $x_c$, a reference view $x^{\mathrm{ref}}$ of the same vehicle with its observed state $\ell^{\mathrm{ref}}$, and $[A, A\odot\ell]$, the lamp map and a state map that marks the lamps to be lit.
Its output $\hat{y}$ is composited only inside the lamp mask, $y=x_c\odot(1-M)+\hat{y}\odot M$.
Each training crop is aligned with a real photo in its observed state $\ell^{\mathrm{obs}}$.
As the editor receives the intention only through $\ell$, we flip $\ell$ with all other inputs fixed and adapt counterfactual fine-tuning~\cite{ribeiro2023counterfactuals}, applying a reconstruction loss to the factual output and a counterfactual loss $\mathcal{L}_{\mathrm{cls}}$ to the flipped one:
\begin{equation}
    \mathcal{L}_{\mathrm{light}}=\underbrace{\mathcal{L}_{1}+\mathcal{L}_{\mathrm{LPIPS}}}_{\ell=\ell^{\mathrm{obs}}}+\underbrace{\lambda_{\mathrm{cls}}\mathcal{L}_{\mathrm{cls}}+\lambda_{\mathrm{col}}\mathcal{L}_{\mathrm{col}}}_{\ell=1-\ell^{\mathrm{obs}}},
\end{equation}
where the $\mathcal{L}_1$ and learned perceptual image patch similarity (LPIPS)~\cite{zhang2018lpips} terms compare $y$ with the photo inside the dilated lamp region, $\mathcal{L}_{\mathrm{cls}}$ requires a frozen ResNet-18 light-state classifier to read the flipped state from $y$, which discourages $\mathcal{E}$ from copying its input, and $\mathcal{L}_{\mathrm{col}}$ penalizes hue drift of lit lamps.

\subsection{Visual-feedback Interaction Learning}
\label{sec:interaction_learning}
IL ignores how the ego reacts to rendered images, which shifts the distribution of interaction features, so we fine-tune the BV policy with MARL~\cite{luo2026multi} in the closed loop.

\noindent\textbf{Closed-loop Markov game with visual feedback.}
We model the closed loop as a Markov game in which the BVs are agents sharing the policy $\pi^{\mathrm{BV}}_\theta$, while the renderer and the ego planner belong to the environment.
At step $t$ ($\Delta t=0.5$\,s), the state $s_t$ holds the map and the recent trajectories and signal light states of the ego and $N$ BVs.
Each BV $i$ plans $\tau^i_t\sim\pi^{\mathrm{BV}}_\theta(\cdot\,|\,o^i_t)$ from its BV-centric state observation $o^i_t$, executes its first $\Delta t$ as action $a^i_t$, and sets its signal light states $\ell^i_t$. The ego observes only the images rendered with enhanced assets and signal lights and responds with $u_t=\pi^{\mathrm{EGO}}(\mathcal{R}(s_t,\ell_t))$, and all vehicles then move by $s_{t+1}=\mathcal{T}(s_t,\{a^i_t\},u_t)$.
The policy $\pi^{\mathrm{BV}}_\theta$ adds to the frozen $\pi^{\mathrm{BV}}_{\mathrm{IL}}$ a residual head $r_\theta$, which samples an offset $\delta$ on $\Delta p$ from a Gaussian with a zero-initialized mean.
For training, we run a group of $G$ episodes of each scene from the same initial state and ego seed, in which the BVs interacting with the ego sample $\delta$ and the others take its mean. 

\noindent\textbf{Distribution-level reward.}
For each interaction phase, from the onset $t_0$ to the end $t_1$ of an acceleration or deceleration, we extract three features $\xi_m$: the time-to-collision at $t_0$ ($\mathrm{TTC}_{t_0}$), the signed peak acceleration during $[t_0, t_1]$ ($a_{\max}$), and the relative speed at $t_1$ ($v^{\mathrm{rel}}_{t_1}$).
Let $q_m$ and $p_m$ be their per-episode histogram counts on training logs and on the episodes of the current iteration.
The return of BV $i$ in episode $g$ favors under-covered interaction bins and penalizes physical violations:
\begin{equation}
    R^i_g=\lambda_d\!\sum_{e\in\mathcal{P}^i_g}\sum_{m}\log\frac{q_m(\xi_m(e))}{p_m(\xi_m(e))}-\sum_k\lambda_k\,C^i_{k,g},
\end{equation}
where $\mathcal{P}^i_g$ are the interaction phases of BV $i$ in the episode and $C^i_{k,g}$ accumulates its collisions, off-road, and progress deviation; the team return is $R_g=\sum_{i=1}^{N}R^i_g$.

\noindent\textbf{Group-relative residual policy optimization.}
Following multi-agent group relative policy optimization (MAGRPO)~\cite{liu2026llm}, all exploring BVs share the group-relative advantage of the team return, $\hat{A}_g=(R_g-\mathrm{mean}_{g'}R_{g'})/\mathrm{std}_{g'}R_{g'}$ over the $G$ episodes $g'$ of the same scene, since a residual also changes other BVs' interactions through the ego response.
As the initial state and ego seed are shared, $R_g$ varies within a group only through the sampled residuals.
The policy is updated with a clipped objective regularized toward the initial policy:
\begin{equation}
    \begin{split}
    \mathcal{L}(\theta)=-\mathbb{E}_{g,\,i\in\mathcal{F},\,t}\Big[&\min\!\big(\rho^i_{g,t}\hat{A}_g,\mathrm{clip}(\rho^i_{g,t},1\!-\!\varepsilon,1\!+\!\varepsilon)\hat{A}_g\big)\\
    &-\beta\,D_{\mathrm{KL}}\big(\pi^{\mathrm{BV}}_\theta(\cdot|o^i_{g,t})\,\|\,\pi^{\mathrm{BV}}_{\theta,0}(\cdot|o^i_{g,t})\big)\Big],
    \end{split}
\end{equation}
where $\mathcal{F}$ are the BVs sampling actions in the group and $\rho^i_{g,t}=\pi^{\mathrm{BV}}_\theta(a^i_{g,t}|o^i_{g,t})/\pi^{\mathrm{BV}}_{\theta_{\mathrm{old}}}(a^i_{g,t}|o^i_{g,t})$ is the probability ratio at re-planning step $t$ of episode $g$.
\section{Experiments}
\label{sec:experiments}

\subsection{Experimental Setup}
\label{sec:setup}
\begin{figure*}[t]
    \centering
    \includegraphics[width=0.95\textwidth]{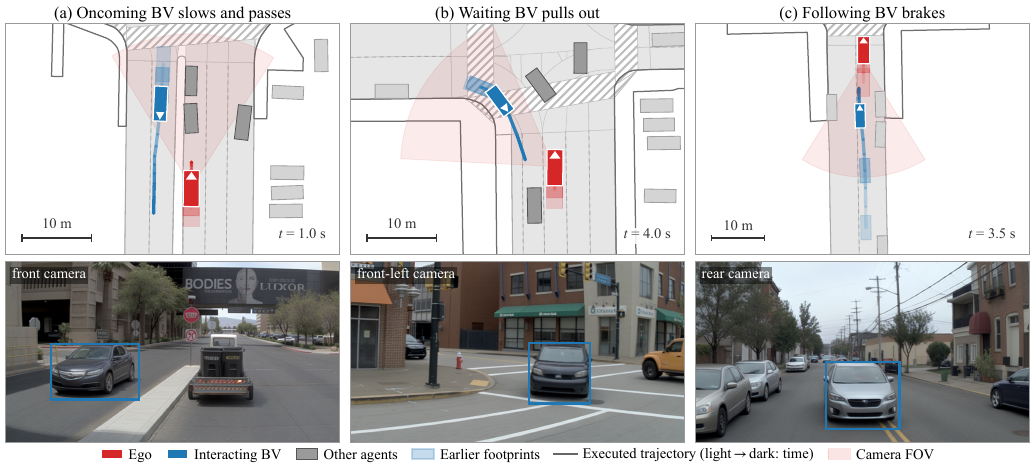}
    \caption{Closed-loop interactions generated by our BV policy. Top: ego-centric bird's-eye view (BEV) with executed trajectories; bottom: our rendering at the marked time from the camera shown in the BEV, with the interacting BV boxed.}
    \label{fig:f1_qual}
\end{figure*}

\noindent\textbf{Closed-loop benchmark.}
We evaluate on the $288$ long-tail \texttt{navtest} failure scenarios~\cite{li2026worldengine} of NAVSIM~\cite{dauner2024navsim} (from $89$ nuPlan~\cite{caesar2021nuplan} logs).
Each episode runs $12$ steps at $0.5$\,s with $8$ surround cameras.
The ego planner is Alpamayo~1.5~\cite{nvidia2025alpamayo}, fed with the front camera only at $2$\,Hz and never fine-tuned.
All methods share the same scenarios, ego planner, enhanced assets, and light signaling, and differ only in the evaluated component.
BV policies are imitation-learned on INTERACTION~\cite{zhan2019interaction} and fine-tuned on nuPlan \texttt{navtrain} scenarios, which have no log overlap with the test set.

\noindent\textbf{Metrics.}
\emph{Interaction realism} compares the distributions of $\mathrm{TTC}_{t_0}$, signed $a_{\max}$, and $v^{\mathrm{rel}}_{t_1}$ between closed-loop rollouts and logged ground truth (GT) over the $1{,}715$ dynamic BVs, using Jensen--Shannon (JS) divergence and histogram overlap, together with the BV collision rate and route progress.
\emph{Novel-view fidelity} is measured by peak signal-to-noise ratio (PSNR), structural similarity index measure (SSIM), and LPIPS on full images and vehicle regions of held-out frames.
\emph{Light signaling realism} is evaluated by the state correctness read out by an independent vision-language model (VLM, Qwen2.5-VL-7B~\cite{bai2025qwen25vl}) never used in training, lamp-region similarity to real photos, and body preservation.
Confidence intervals (CIs) are obtained by scene- or vehicle-level paired bootstrap.

\subsection{Interaction Realism}
\label{sec:exp_f1}
\begin{figure*}[t]
    \centering
    \includegraphics[width=\textwidth]{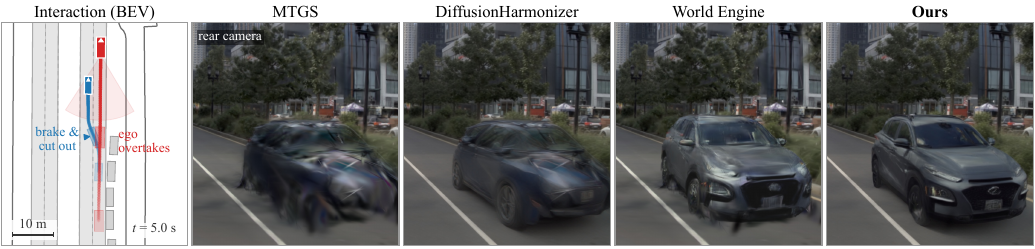}
    \caption{A novel view exposed by interaction: the rear camera of the ego (red) sees the front of an overtaken BV (blue). MTGS smears the body, which DiffusionHarmonizer cannot restore; World Engine has frayed edges; ours is complete and keeps the observed appearance.}
    \label{fig:f2_inter}
\end{figure*}

We compare with WMs that claim realistic interactions between the ego and reactive BVs: RealEngine~\cite{jiang2025realengine} (BVs driven by DiffusionDrive~\cite{liao2025diffusiondrive} planners), OmniNWM~\cite{li2026omninwm}, and World Engine~\cite{li2026worldengine} (SOTA).

\begin{table}[t]
\centering
\caption{Interaction realism on $288$ closed-loop scenarios. GT: logged reference on the same BVs and timestamps; its collisions are overlaps of logged boxes. $^\dagger$BV agents run in our closed loop and renderer. $^\ddagger$BV states estimated from generated videos.}
\label{tab:f1_dist}
\small
\setlength{\tabcolsep}{2.5pt}
\resizebox{\linewidth}{!}{%
\begin{tabular}{lccccccc}
\toprule
\multirow{2}{*}{Method} & \multicolumn{4}{c}{Overlap $\uparrow$} & \multirow{2}{*}{\makecell{Mean\\JS $\downarrow$}} & \multirow{2}{*}{\makecell{Collision\\rate $\downarrow$}} & \multirow{2}{*}{\makecell{Progress\\(m) $\uparrow$}} \\
\cmidrule(lr){2-5}
 & $\mathrm{TTC}_{t_0}$ & $a_{\max}$ & $v^{\mathrm{rel}}_{t_1}$ & Mean & & & \\
\midrule
GT (log) & -- & -- & -- & -- & -- & 3.50\% & 27.66 \\
\midrule
RealEngine$^\dagger$~\cite{jiang2025realengine} & 0.8588 & 0.4324 & 0.6778 & 0.6563 & 0.1152 & 12.71\% & 20.30 \\
OmniNWM$^\ddagger$~\cite{li2026omninwm} & 0.7455 & 0.1492 & 0.8544 & 0.5830 & 0.1886 & 23.08\% & 14.65 \\
World Engine~\cite{li2026worldengine} & 0.9527 & 0.5771 & 0.8498 & 0.7932 & 0.0485 & 10.38\% & 25.91 \\
\textbf{Ours} & \textbf{0.9775} & \textbf{0.8719} & \textbf{0.9102} & \textbf{0.9198} & \textbf{0.0074} & \textbf{9.74\%} & \textbf{28.10} \\
\bottomrule
\end{tabular}}
\end{table}

In \cref{tab:f1_dist}, our BVs are the closest to real-world driving logs on all three features, reducing the mean JS divergence of World Engine by $85\%$.
For all compared WMs, $a_{\max}$ is the worst-matched feature (overlap at most $0.58$), whereas $\mathrm{TTC}_{t_0}$, \ie, when a reaction starts, is matched far better (up to $0.95$): existing BV models react at about the right time but not in the right way.
Their errors even go in opposite directions: the BVs of World Engine react about twice as hard as the logged vehicles (median $|a_{\max}|$ of $4.7$ \vs $2.3$\,m/s$^2$), while those of RealEngine decelerate in $73\%$ of the interaction phases, against $46\%$ in the logs, and travel $27\%$ less.
Our BVs keep the route progress within $2\%$ of the logs at the lowest collision rate without trading progress for safety, and pass, wait, or brake depending on the ego's motion (\cref{fig:f1_qual}).

\subsection{Novel-view Fidelity}
\label{sec:exp_f2}
\noindent\textbf{Protocol.}
Since simulated interactions have no GT, we hold out all camera images and LiDAR at selected timestamps of $11$ nuPlan logs and render the full scenes at the real camera and vehicle poses of these frames; no held-out frame is used for training or model selection.
We evaluate $1{,}184$ held-out images and $365$ vehicle regions, averaging within each log and then across logs.
The methods compared include reconstruction (MTGS), diffusion-based enhancement (Difix3D+, DiffusionHarmonizer), and whole-asset generation (Asset Harvester, World Engine).

\begin{table}[t]
\centering
\caption{Novel-view fidelity on held-out frames of $11$ logs. All methods share the MTGS background and differ in the target vehicle assets or post-processing.}
\label{tab:f2_main}
\small
\setlength{\tabcolsep}{2.5pt}
\resizebox{\linewidth}{!}{%
\begin{tabular}{lcccccc}
\toprule
\multirow{2}{*}{Method} & \multicolumn{3}{c}{Vehicle region} & \multicolumn{3}{c}{Full image} \\
\cmidrule(lr){2-4}\cmidrule(lr){5-7}
 & PSNR$\uparrow$ & SSIM$\uparrow$ & LPIPS$\downarrow$ & PSNR$\uparrow$ & SSIM$\uparrow$ & LPIPS$\downarrow$ \\
\midrule
MTGS~\cite{chen2025mtgs} & 17.05 & 0.411 & 0.466 & 18.17 & 0.571 & 0.468 \\
Difix3D+~\cite{wu2025difix3d} & 16.91 & 0.396 & \textbf{0.415} & 18.18 & 0.539 & \textbf{0.383} \\
DiffusionHarmonizer~\cite{diffusionharmonizer2026} & 17.11 & 0.406 & 0.432 & 18.08 & 0.560 & 0.427 \\
Asset Harvester~\cite{assetharvester2026} & 15.24 & 0.359 & 0.504 & 18.06 & 0.570 & 0.469 \\
World Engine~\cite{li2026worldengine} & 15.97 & 0.375 & 0.490 & 18.08 & 0.570 & 0.469 \\
\textbf{Ours} & \textbf{17.97} & \textbf{0.438} & 0.448 & \textbf{19.01} & \textbf{0.584} & 0.432 \\
\bottomrule
\end{tabular}}
\end{table}

\noindent\textbf{Results and analysis.}
In \cref{tab:f2_main}, our method achieves the best PSNR and SSIM on both vehicle regions and full images, with a vehicle PSNR $2.0$\,dB above World Engine.
Notably, replacing whole vehicles with generated assets (Asset Harvester and World Engine) falls $1.1$--$1.8$\,dB \emph{below} the unenhanced MTGS reconstruction: a generated asset is plausible, but it overwrites the regions observed in the log with content that no longer belongs to the logged vehicle.
Likewise, whole-frame diffusion enhancers obtain the lowest LPIPS by sharpening textures, yet reduce vehicle SSIM and hardly change PSNR, since the added details are not those of the real vehicle.
Generation thus helps only where observations are missing, which is why we repair selectively and bound the image-space correction.
Our LPIPS is not the best for the same reason: we favor fidelity to the logged vehicle over sharpness (\cref{tab:f2_abl}).
\cref{fig:f2_inter} shows the opposite regime, a side never observed in the log, where only ours is complete and keeps the observed appearance.

\subsection{Vehicle Light Signaling Realism}
\label{sec:exp_f3}
\begin{figure*}[t]
    \centering
    \includegraphics[width=0.95\textwidth]{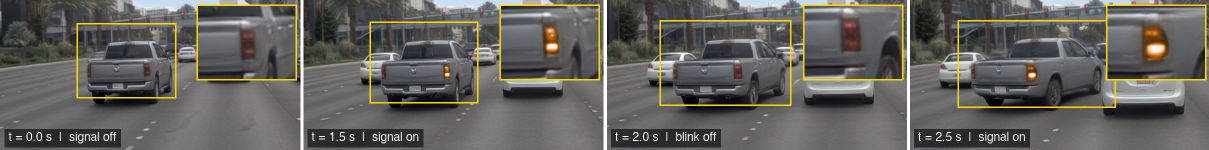}
    \caption{Rendered turn signal during a closed-loop cut-in: it turns on at $t=1.5$\,s, before visible lateral motion, and blinks at $1$\,Hz.}
    \label{fig:f3_seq}
\end{figure*}

We compare with the original rendering and with instruction-based editors InstructPix2Pix~\cite{brooks2023instructpix2pix}, MagicBrush~\cite{zhang2023magicbrush}, and FLUX~\cite{labs2025flux}, all applied to the same rendered vehicle crops.

\begin{figure}[t]
    \centering
    \includegraphics[width=\linewidth]{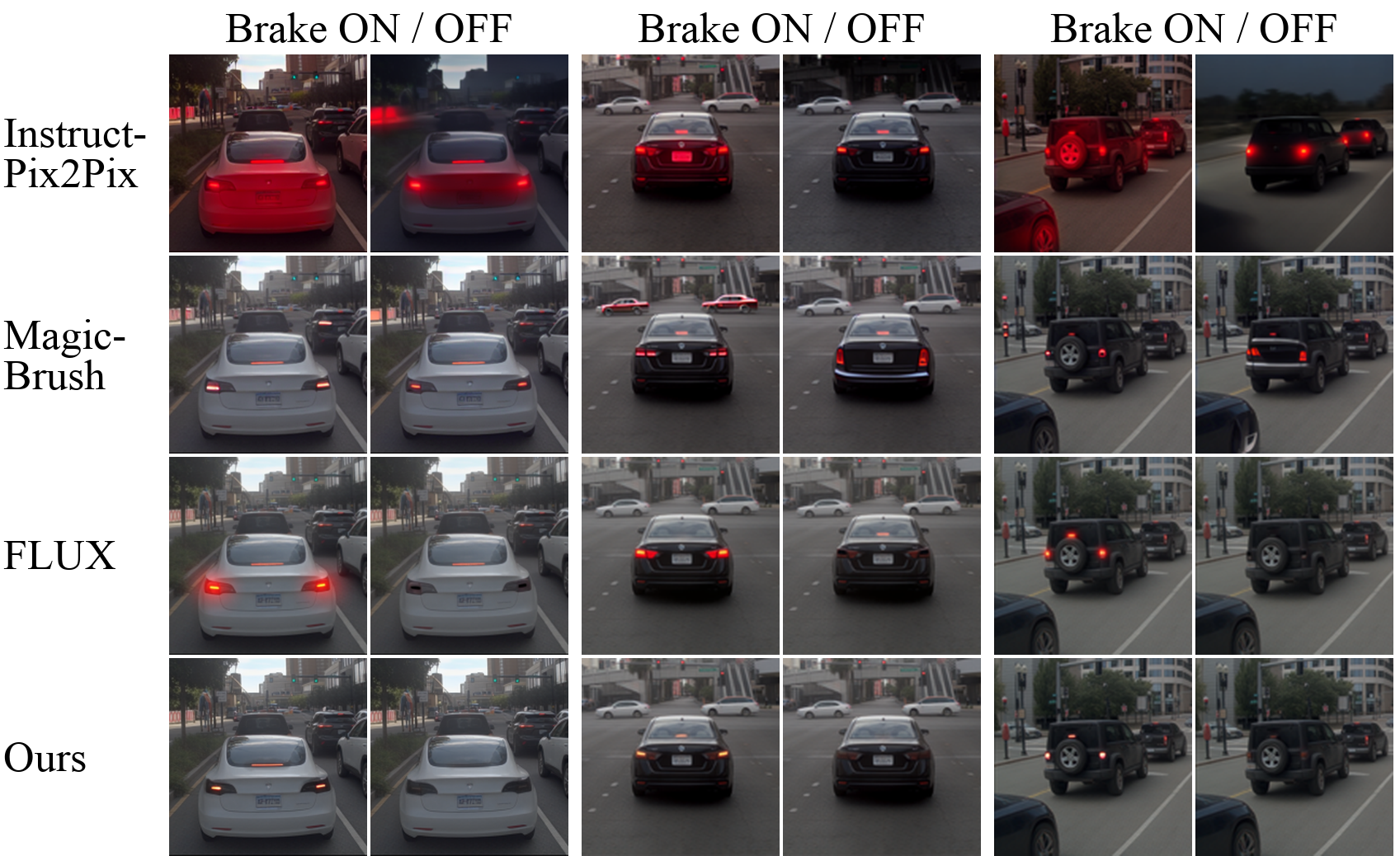}
    \caption{Brake-light editing on rendered vehicles (each pair: target ON / OFF). General editors alter the body or the scene and often fail to switch lights off; ours edits only the anchored lamp regions.}
    \label{fig:f3_qual}
\end{figure}

\begin{table}[t]
\centering
\caption{Signal light state correctness (VLM readout). Brake: bidirectional counterfactual edits ($435$ items, $115$ vehicles); turn: left/right activation ($272$ items, $136$ vehicles). mF1: macro-F1.}
\label{tab:f3_state}
\small
\setlength{\tabcolsep}{3pt}
\resizebox{\linewidth}{!}{%
\begin{tabular}{lccccc}
\toprule
Method &
\makecell{Brake\\mF1 $\uparrow$} &
\makecell{Brake\\On/Off rec. $\uparrow$} &
\makecell{Turn\\mF1 $\uparrow$} &
\makecell{Turn\\On rec. $\uparrow$} &
\makecell{Turn\\Wrong/Miss $\downarrow$} \\
\midrule
Original rendering & 0.400 & 29.7\% / 51.9\% & -- & -- & -- \\
InstructPix2Pix~\cite{brooks2023instructpix2pix} & 0.442 & 90.9\% / 13.9\% & 0.393 & 88.2\% & 47.4\% / 11.8\% \\
MagicBrush~\cite{zhang2023magicbrush} & 0.525 & 81.7\% / 29.6\% & 0.347 & 68.0\% & 35.7\% / 32.0\% \\
FLUX~\cite{labs2025flux} & 0.830 & 94.5\% / 71.8\% & 0.685 & 91.5\% & 25.4\% / 8.5\% \\
\textbf{Ours} & \textbf{0.839} & \textbf{95.0\%} / \textbf{75.5\%} & \textbf{0.769} & \textbf{94.9\%} & \textbf{19.5\%} / \textbf{5.1\%} \\
\bottomrule
\end{tabular}}
\end{table}

\begin{table}[t]
\centering
\caption{Lamp-region similarity to real photos and body change outside lamps (same-state / target-state references).}
\label{tab:f3_photo}
\small
\setlength{\tabcolsep}{3pt}
\resizebox{\linewidth}{!}{%
\begin{tabular}{lccc}
\toprule
Method & Lamp LPIPS $\downarrow$ & Lamp PSNR $\uparrow$ & Body changed $\downarrow$ \\
\midrule
Original rendering & 0.322 / 0.454 & 18.45 / 16.43 & -- \\
InstructPix2Pix~\cite{brooks2023instructpix2pix} & 0.578 / 0.598 & 11.40 / 11.40 & 81.13\% / 87.61\% \\
MagicBrush~\cite{zhang2023magicbrush} & 0.456 / 0.533 & 14.48 / 13.41 & 12.06\% / 13.30\% \\
FLUX~\cite{labs2025flux} & 0.399 / 0.507 & 16.09 / 14.32 & 15.01\% / 4.36\% \\
\textbf{Ours} & \textbf{0.287} / \textbf{0.400} & \textbf{18.55} / \textbf{17.10} & \textbf{0.14\%} / \textbf{0.06\%} \\
\bottomrule
\end{tabular}}
\end{table}

\noindent\textbf{State correctness and locality.}
In \cref{tab:f3_state}, our method is the most accurate on both brake lights and turn signals, with a turn-signal mF1 $12\%$ above FLUX, although the target state was never observed on many test vehicles.
Switching a light off is harder than switching it on for every editor: general editors easily add a glow but rarely restore an unlit lamp (off-recall as low as $13.9\%$), and deactivation remains our main failure mode as well.
FLUX is close to ours on brake lights, but state correctness alone is insufficient for a WM: general editors repaint $4\%$--$88\%$ of the body pixels (\cref{tab:f3_photo,fig:f3_qual}), so a vehicle would change its appearance whenever its lights switch.
Compositing only inside the 3D-anchored lamp regions, ours changes less than $0.2\%$ of the body and is closest to real photos of the target state.

\noindent\textbf{Effect on the ego.}
In \cref{fig:f3_seq}, the turn signal is activated before the visible lateral motion, showing the intention before the action.
To test whether such signals matter to the ego, we run the $288$ scenarios with and without light rendering under otherwise identical settings.
The ego trajectory deviates by more than $10$\,cm in $24.3\%$ / $21.5\%$ of the scenarios for Alpamayo~1.5 / openpilot~\cite{goff2025openpilot}.
The response is concentrated: for Alpamayo~1.5, rendered lights change the planner's front-view input images in $36\%$ of the scenarios, and the trajectory deviates in $67\%$ of them, by $0.77$\,m on average.
Signal lights are thus a functional input of end-to-end planners, and a WM with static lights changes the tested ego behavior in about a quarter of long-tail scenarios.

\subsection{Ablation Studies}
\label{sec:ablation}

\noindent\textbf{Visual-feedback interaction learning.}
Deploying the IL policy $\pi^{\mathrm{BV}}_{\mathrm{IL}}$ without closed-loop fine-tuning increases the mean JS divergence by $77\%$ (paired difference $+0.0057$, 95\% CI $[+0.0021,+0.0092]$).
IL already matches when BVs start to react ($\mathrm{TTC}_{t_0}$ overlap $0.96$) but not how: facing an ego that plans from rendered images, its BVs over-brake, decelerating in $66\%$ of the interaction phases against $46\%$ in the logs, and travel $9\%$ less.
Closed-loop fine-tuning reduces this share to $50\%$ and restores the route progress; such a shift~\cite{ross2011dagger,zhang2025catk} appears only in the closed loop and is not corrected by imitation alone.

\begin{table}[t]
\centering
\caption{Ablation of interaction-guided asset enhancement (vehicle region). Rows are cumulative except the indented control.}
\label{tab:f2_abl}
\small
\setlength{\tabcolsep}{4pt}
\begin{tabular*}{\linewidth}{@{\extracolsep{\fill}}lccc}
\toprule
Variant & PSNR$\uparrow$ & SSIM$\uparrow$ & LPIPS$\downarrow$ \\
\midrule
Native MTGS asset & 17.05 & 0.411 & 0.466 \\
+ diagnosis-guided repair & 17.59 & 0.421 & 0.464 \\
+ calibration (i) & 17.62 & 0.423 & 0.463 \\
\quad calib.\ + harmonizer only & 17.55 & 0.413 & \textbf{0.430} \\
+ fusion, fixed weights & 17.80 & 0.431 & 0.458 \\
+ fusion, learned weights (ii, full) & \textbf{17.97} & \textbf{0.438} & 0.448 \\
\bottomrule
\end{tabular*}
\end{table}

\noindent\textbf{Interaction-guided asset enhancement.}
In \cref{tab:f2_abl}, diagnosis-guided repair is the largest single step, contributing $0.54$\,dB, nearly $60\%$ of the total gain.
Replacing the whole vehicle with the same donor instead loses $1.8$\,dB (\cref{tab:f2_main}), so the gain comes from selecting where to repair rather than from the generator.
Repair raises PSNR and SSIM but hardly LPIPS; the perceptual gain comes from the harmonizer, of which bounded fusion keeps about half without the PSNR drop of accepting it everywhere.

\begin{table}[t]
\centering
\caption{Ablation of light signaling. On-recall: vehicles observed only with signal lights off ($2{,}811$ items). Jump rate: lamp-center shift above $5\%$ of the box size between consecutive steps.}
\label{tab:f3_abl}
\small
\setlength{\tabcolsep}{0pt}
\begin{tabular*}{\linewidth}{@{\extracolsep{\fill}}lccc}
\toprule
Variant & Brake mF1 $\uparrow$ & On-recall $\uparrow$ & Jump rate $\downarrow$ \\
\midrule
w/o counterfactual loss & 0.626 & 35.4\% & -- \\
w/o 3D anchors (2D seg.) & \textbf{0.845} & 83.7\% & 41.4\% \\
w/o reference view & 0.825 & 85.1\% & -- \\
w/o state map & 0.813 & 84.1\% & -- \\
\textbf{Full} & 0.839 & \textbf{86.1\%} & \textbf{31.9\%} \\
\bottomrule
\end{tabular*}
\end{table}

\noindent\textbf{Light signaling.}
In \cref{tab:f3_abl}, the counterfactual loss is the key to state switching: without it, only observed states are supervised, and the editor turns on the lights of vehicles observed only with lights off in about a third of the cases instead of $86\%$.
3D anchors do not improve single-frame accuracy; we adopt them for the temporal stability that a closed loop requires, as the lamp-center jump rate rises by $9.5$ points (95\% CI $[+5.0,+14.3]$) without them.
The reference view and the state map have only minor gains.

\section{Conclusion}
\label{sec:conclusion}
We present a closed-loop interactive WM that couples behavior modeling and visual rendering.
Compared with the SOTA WM, it reduces the divergence from real-world interaction behaviors by $85\%$ on long-tail closed-loop scenarios and improves novel-view vehicle PSNR by $2.0$\,dB, and its vehicle light signaling is consistent with driving intentions.
Our experiments further lead to three findings.
(i) BV behaviors in existing WMs differ from real driving in how vehicles react rather than when. Without closed-loop fine-tuning, BVs over-brake when facing an ego that plans from rendered images.
(ii) Generation should complement rather than replace observation: replacing whole vehicle assets with generated ones lowers fidelity below the unenhanced reconstruction.
(iii) Signal lights are a functional input of end-to-end planners. When rendered lights change the planner's input images, the ego trajectory deviates in two thirds of the cases. A WM with static lights cannot elicit this response.
In future work, we will use our WM to accelerate closed-loop training: BVs optimized against a tested planner can automatically generate long-tail scenarios tailored to its weaknesses.

\section*{LLM Usage Statement}
\label{sec:llm}
We used LLMs (Claude, ChatGPT, and Gemini) for polishing the text, literature discovery, and generating code snippets.
The research problem, the method, and the experimental design were conceived by the authors, who also conducted and analyzed the experiments.
All LLM-assisted text and code and all cited works were reviewed and verified by the authors, who take full responsibility for the content.

{
    \small
    \bibliographystyle{ieeenat_fullname}
    \bibliography{main}
}

\end{document}